\documentclass{article}

\usepackage{PRIMEarxiv}

\usepackage[utf8]{inputenc} 
\usepackage[T1]{fontenc}    
\usepackage{hyperref}       
\usepackage{url}            
\usepackage{booktabs}       
\usepackage{amsfonts}       
\usepackage{nicefrac}       
\usepackage{microtype}      
\usepackage{lipsum}
\usepackage{graphicx}
\usepackage{amsmath}
\usepackage{algorithm2e}
\usepackage{comment}
\usepackage{color,soul}
\graphicspath{{media/}}     
\usepackage{xcolor}         
\usepackage{graphicx}  
\usepackage{wrapfig}   
\usepackage[numbers]{natbib} 

\title{Superpixel Boundary Correction for Weakly-Supervised Semantic Segmentation on Histopathology Images
}

\author{
  Hongyi Wu \\
  School of Management \\
  University of Science and Technology of China \\
  \texttt{ahwhy@mail.ustc.edu.cn} \\
   \And
  *Hong Zhang \\
  School of Management \\
  University of Science and Technology of China \\
  \texttt{zhangh@ustc.edu.cn} \\
}

\begin{document}
\maketitle

\begin{abstract}
With the rapid advancement of deep learning, computational pathology has made significant progress in cancer diagnosis and subtyping. Tissue segmentation is a core challenge, essential for prognosis and treatment decisions. Weakly supervised semantic segmentation (WSSS) reduces the annotation requirement by using image-level labels instead of pixel-level ones. However, Class Activation Map (CAM)-based methods still suffer from low spatial resolution and unclear boundaries. To address these issues, we propose a multi-level superpixel correction algorithm that refines CAM boundaries using superpixel clustering and floodfill. Experimental results show that our method achieves great performance on breast cancer segmentation dataset with mIoU of 71.08$\%$, significantly improving tumor microenvironment boundary delineation.
\end{abstract}


\section{Introduction}

The progress in deep learning has significantly advanced computational pathology, reducing pathologists from repetitive tasks such as cancer diagnosis and subtyping. One of the core challenges in this field is automated tissue segmentation, which is crucial since tumor microenvironment (TME) plays a key role in tumor growth and development. Accurate segmentation of TME in histopathology can provide valuable insights into patient prognosis. This will support oncologists in treatment decisions, clinical therapies and effective cancer care.

Currently, most deep learning methods for TME segmentation tasks are fully supervised. Due to the extremely large size of histopathology images (with hundreds of thousands of pixels) and the limited availability of pathologists, fully supervised annotations for whole slide images (WSIs) are an expensive and challenging task to implement. Reducing the amount of annotation has become a critical issue,  often addressed through active learning~\citep{yang2017suggestive}, semi-supervised learning~\citep{liu2020semi}, and weakly supervised learning~\citep{chan2021comprehensive}. Weakly supervised semantic segmentation (WSSS) only requires image-level labels instead of pixel-level annotations, thus saving significant costs for segmentation tasks of WSIs ~\citep{han2022wsss4luad}.

For Convolutional Neural Networks (CNN)-based methods, class activation map (CAM)  ~\citep{zhou2016learning} is a technique used to highlight the regions of an image that are most relevant to a specific class prediction. However, The convenience of CAM highlighting high-contribution regions can potentially lead to a loss in efficiency. Drawbacks of CAM are mostly caused by low spatial resolution and lack of boundary awareness. First, in many cases, the CAM heatmap is based on the output of high-level layers of the neural network, with lower spatial resolution at the same time. This can cause the boundary of the object to appear blurred or fuzzy, making it difficult to locate the semantic shape precisely. Moreover, since CAM is primarily concerned with classification and the most discriminative regions, it may not cover the full spatial extent of the object, especially its edges or boundaries, where less distinctive features are present. This results in incomplete or imprecise boundary delineation.

\begin{figure}[htbp]
  \centering
  \includegraphics[width=0.6\textwidth]{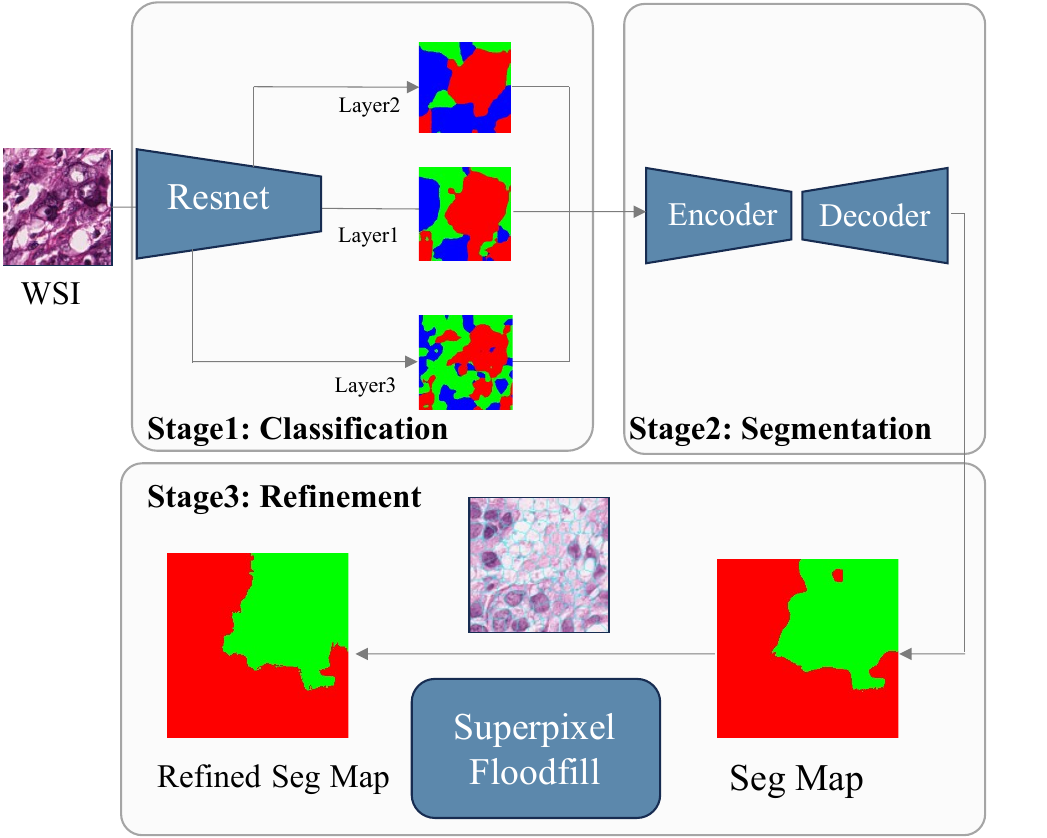}
  \hspace{-10pt}
  \caption{Pipeline of multi-layer superpixel correction for WSSS on WSIs.}
  \label{fig:pipeline}
\end{figure}

Hence, the intractability in CAM for WSIs can be described as a boundary problem. To address this issue, we introduce the concept of superpixels in images, as superpixel clustering can reflect natural boundaries of an image, which is especially important for pathological segmentation. With siperpixels we refine the boundaries of CAM-supervised segmentation using floodfill (Fig. ~\ref{fig:pipeline}). We also address the boundary scale problem by merging CAMs of different depths.


The main contributions of this paper are as follows. First, we propose a superpixel correction algorithm to refine the boundary of CAM. It combines superpixel clustering into a multi-layer segmenation backbone to improve the quality of segmentation mask. Second, various experiments demonstrate that our algorithm achieves state of the art (SOTA), specificially on BCSS segmentation dataset. Accordingly, our paper further improves the efficiency of WSSS, making pathological image segmentation more lightweight, and thereby facilitating the practical application of computer-assisted healthcare.


\section{Related works}
\label{sec:related}
\subsection{Whole Slide Images}
Pathological images are an important diagnostic basis in medicine, allowing doctors to assess a patient’s cancer status based on tumor microenvironments within these images. With advancements in deep learning, research on pathological images, e.g. WSIs, has progressed more effectively. 
Computational tasks involving WSIs are mainly divided into three domains: image segmentation, image classification, and object detection ~\citep{shen2017deep}. Among these tasks, image segmentation is the most comprehensive and complex task, requiring pixel-by-pixel semantic segmentation to obtain a complete semantic map for observation. Since WSIs often contain hundreds of thousands of pixels, obtaining pixel-level annotations is very costly ~\citep{hanna2020whole}. Therefore, weakly supervised semantic segmentation has shown great advantages, as it only requires image-level classification labels rather than pixel-level annotations.

\subsection{WSSS}
Many WSSS methods utilize Class Activation Maps to obtain pseudo-masks. CAMs are then expanded, refined, and augmented to produce fine-grained segmentation masks. The development of WSSS with CAMs follows two lines: improving the generation of pseudo-masks and improving the expansion of pseoudo-masks. Concentrating on WSIs, some recent methods have progressed in CAM-based approaches. Histo-Seg ~\citep{chan2019histosegnet} trains a CNN on the gradient-weighted CAM and adds post-processing segmentation with a fully-connected conditional random field. SEAM ~\citep{du2022weakly} runs on different views in order to impose regularization of semantic consistency of features between views and facilitate the compactness between classes of the feature space.  C-CAM~\citep{chen2022c} proposes a causal CAM with two cause-effect chains including category-causality chain and anatomy-causality chain.
WSSS-Tissue ~\citep{han2022multi} is a method utilizing multi-layer pseudo-supervision with progressive dropout attention. PistoSeg ~\citep{fang2023weakly} attains pseudo-masks with a synthesis of Mosaic transformation on raw weakly-supervised datasets.

\section{Method}

\subsection{Multi-Layer Pseudo-Mask Fusion}
\label{sec: multi}

The organization of tissues is random and dispersed, which means a single image patch may contain more than one tissue type. Therefore, we consider tissue segmentation as a multi-label classification problem.

Some variable notations are as follows:
\begin{itemize}
    \item \( X \in \mathbb{R}^{H \times W} \): raw input data of an \( H \times W \) pixel matrix where H is its height and W is its width;
    \item \( Y \in \mathbb{R}^k \):  image classification label vector of \( k \)-dimension;
    \item \( D_{\text{train}} = \{(x_i, y_i), i=1,\ldots,n_{\text{train}}\} \): a training dataset, where $(x_i,y_i)_{i=1}^{n_{{\text{train}}}}$ are i.i.d. copies of $(X,Y)$ and $n_{{\text{train}}}$ is its size;
    \item \( D_{\text{test}} = \{(x_i', y_i'), i=1,\ldots,n_{\text{test}}\} \): a testing dataset where $(x_i',y_i')_{i=1}^{n_{\text{test}}}$ are i.i.d. copies of $(X,Y)$ and $n_{{\text{test}}}$ is its size.
\end{itemize}

Considering a CNN-based classification model $f_{\text{cls}}$ with a weight matrix $W_{\text{cls}}$, the training process is represented by $\hat{y} = f_{\text{cls}}(x, W_{\text{cls}})$. In this process, raw image features undergo convolutional layers to transform into deep feature maps. Function 
$T: f_{\text{cls}}(x, W_{\text{cls}}) \rightarrow m$  represents the extraction of final feature map $m$ from the network. After that, the classification logit $\hat{z}_c$ for class $c$ can be obtained using a global weighted average pooling layer:
$$
\hat{z}_c = \sum_{k} w_{c,k} \text{GAP}_k(m_k),
$$
where $\text{GAP}_k$ is global average pooling layer, $w_{c,k}$ is weight of class $c$ to feature map $m_k$ in linear layer.
After training the above classification model, we can generate a pseudo mask $ p $ (an $H\times  W$ matrix) through class activation mapping, with the  $(i,j)$ element $p(i,j)$ defined as::
$$
p(i,j) = \underset{c}{\text{argmax}} \,\sum_{k} w_{c,k} m_k(i,j). 
$$

With generated pseudo-masks, the training set can be expanded to $ \tilde{D}_{\text{train}} = \{(x_i, y_i,p_i), i=1,\ldots,n_{\text{train}}\} $. On \( \tilde{D}_{\text{train}} \), we can train a pseudo supervised model \( f_{\text{seg}} \) to obtain the final segmentation output \( s \) with weights $W_{seg}$:
$$s = f_{\text{seg}}(x, p, W_{\text{seg}})
$$

Through this pseudo-mask approach, we achieve an approximation under weak supervision. However, due to the discrepancy between pixel-level labels and image-level labels, the spatial information learned simply from the classification network remains incomplete. To bridge this gap, we need to obtain deeper-level image information.

With deep layers, CNN models are capable of learning information at different scales. We can extract pseudo-masks from CAM at various CNN depths and incorporate them into the loss function to reflect multi-scale information ~\citep{han2022multi}. To address this, we select feature maps from three different CNN depths, obtaining three pseudo masks \( p_1, p_2, p_3 \) through CAM. The adjusted loss function is:
\[
\mathcal{L} = \lambda_1 l(s, p_1) + \lambda_2 l(s, p_2) + \lambda_3 l(s, p_3),
\]
where \( l_1, l_2, l_3 \) are cross-entropy losses, and \( \lambda_1, \lambda_2, \lambda_3 \) are tuning parameters.

\subsection{Superpixel Floodfill Refinement}
Due to the lack of pixel-level annotations in weakly supervised learning, the quality of the pseudo-masks largely determines the quality of the segmentation results. As pseudo-masks only represent the areas with the highest predicted probability density and does not restore finer semantic features, relying solely on coarse CAM probability heatmaps  may lead to discrepancies between the image boundaries and the ground truth (GT).

In multi-layer models, the granularity of CAMs generated by GAP of different depths exhibits significant variability. This variability in granularity is primarily reflected in the scale of the boundaries. However, in most regions of primary pseudo masks, the depth gap does not result in noticeable differences in granularity. Meanwhile, we observe that superpixels of CAMs can effectively correct the semantic boundaries, leveraging their inherent properties.

\begin{wrapfigure}[11]{r}{0.5\textwidth}
    \centering
    {\includegraphics[width=0.5\textwidth]{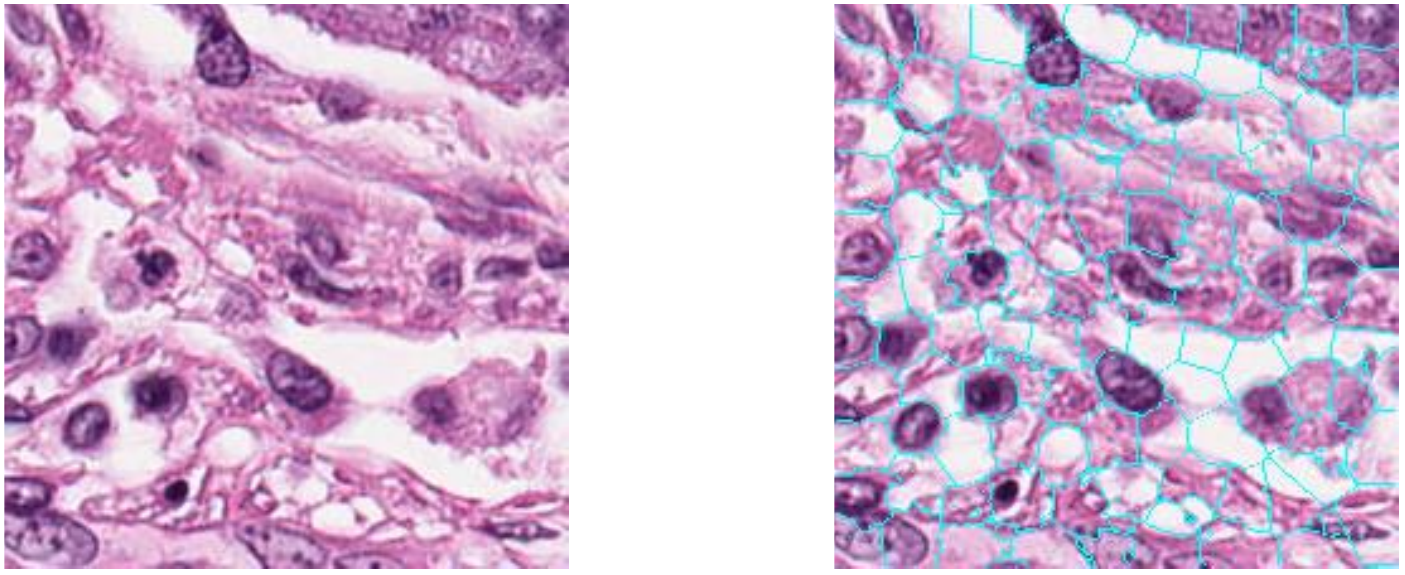}}
    \caption{Illustration of superpixels on WSI}
    \label{sp-illu}
\end{wrapfigure}

Therefore, a fundamental idea is to cluster on basic features (such as color and texture) of the image, thereby partitioning the entire image $x$ into several superpixels $C=\{c_1,c_2,\ldots\}$ (as shown in Fig. ~\ref{sp-illu}), and then making corrections to the pseudo-mask $p$ on this superpixel partition $C$.

We utilize Simple Linear Iterative Clustering (SLIC) algorithm to partition superpixels. We control the SLIC algorithm through three parameters: average cluster size $S$ and compactness $m$. In SLIC, the distance $D$ of pixel vector space is defined for clustering:
$$
D = \sqrt{\left(\frac{\Delta x}{S}\right)^2 + \left(\frac{\Delta y}{S}\right)^2 + \left(\frac{\Delta I}{m}\right)^2},
$$
where $x$ and $y$ represent the coordinate of a pixel point while  
$I$ represents the color intensity. As $m$ increases, the influence of color intensity on distance decreases, and the effect of spatial features on distance becomes more prominent, resulting in smoother superpixel boundaries. Conversely, when $m$ is smaller, the boundary has finer granularity.  

Superpixels from SLIC are not directly related to semantic segmentation but have natural boundaries among them. Observations reveal that many superpixels actually belong to the same category, but predicted to multiple categories due to the limitation of CAMs. To address this, we process the pixel points in the pseudo-mask $p$ with ``floodfill'' corresponding to each superpixel. If a superpixel contains a ``dominant category'' (which means most pixels in this superipixel belong to one category), we assign all pixels in this superpixel to the dominant category. If no dominant category exists, the superpixel is left unchanged. This superpixel floodfill algorithm is detailed in Algorithm. \ref{sfa}.

\begin{algorithm}[H]
\SetAlgoLined
\KwResult{$\{p_{refined}\}$}
\KwData{$(x, y, p)$ $\in \tilde{D}_{train}$}
 $C=\{c_1,c_2,...\}=\text{SLIC}(x)$; 
 
 $K: \text{number of classes}; \tau: \text{threshold parameter}$;

 \For{$c_i$ in C }{
  \For{j in 1:K}{
 $n_{ij} = \left|\{pixel\in c_i, class_p(pixel)=j\}\right|$}\
  $l_i$ = $\underset{j}{{\arg\max}} \frac{|n_{ij}|}{|c_i|}$ \
  
  $r_i=\frac{|n_{i l_{i}}|}{|c_i|}$
  
  \eIf{$r_i > \tau$}{
   $class_{p_{\text{refined}}}(pixel):=l_i, pixel\in c_i$\;
   }{
   continue\;
  }
 }
 \caption{Superpixel Floodfill Algorithm}
 \label{sfa}
\end{algorithm}

\section{Experiments}

\subsection{Dataset and Methods}

\begin{figure}[htbp]
  \centering
  \includegraphics[width=1\textwidth]{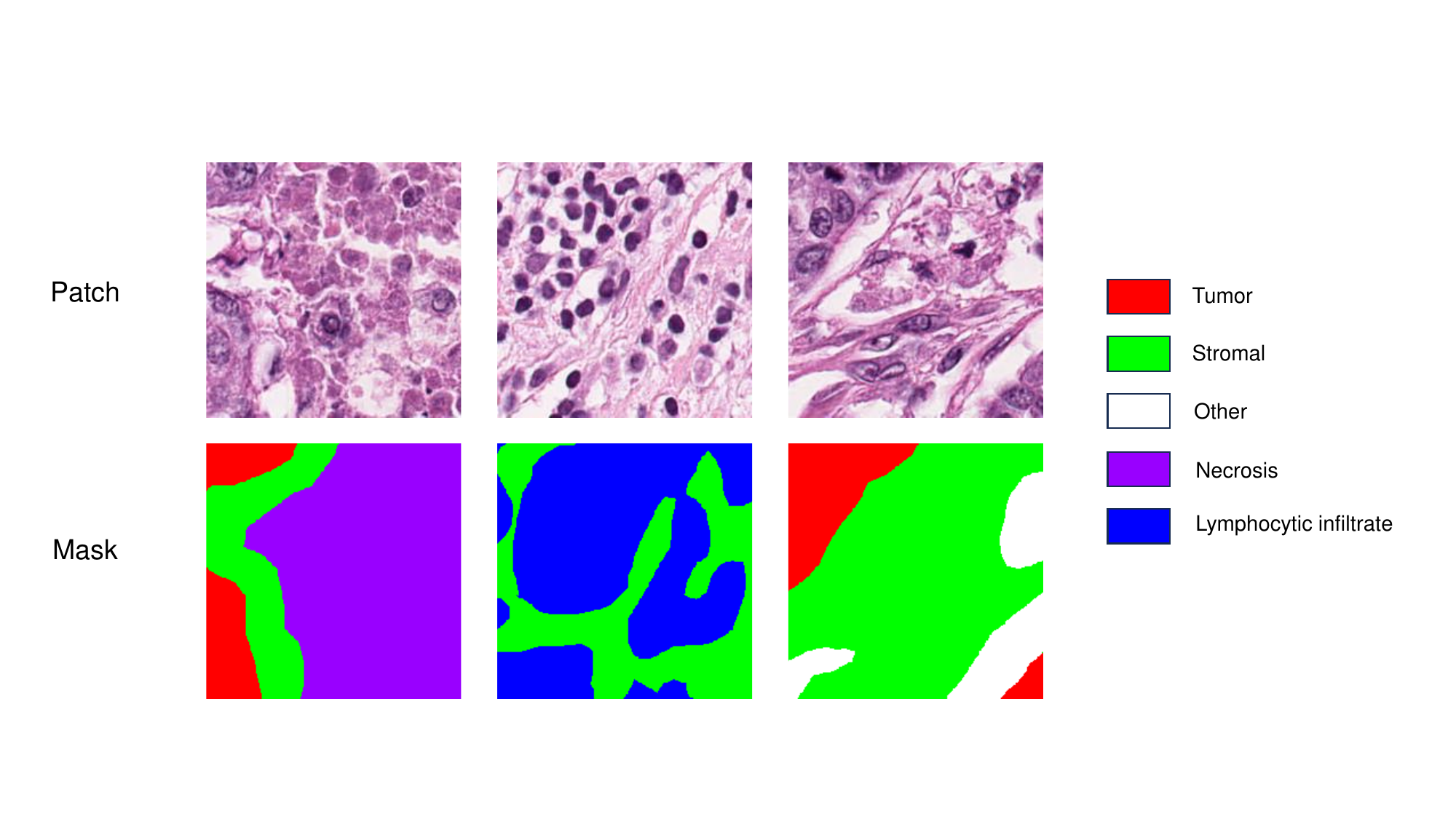}
  \hspace{-10pt}
  \caption{Illustration of patches and masks in BCSS dataset.}
  \label{fig:bcss}
\end{figure}

We illustrate the proposed algorithm with the Breast Cancer Semantic Segmentation (BCSS) dataset, which consists of 151 representative regions of interest (ROIs), i.e., H\&E-stained whole slide images of breast cancer. Experienced pathologists classify the images without providing pixel-level mask information. These 151 whole slide images are cut into smaller slices of 224 × 224 pixels, resulting in 23,422 training images, 3,418 validation images, and 4,986 test images. As shown in Fig. ~\ref{fig:bcss}, the BCSS dataset contains four types of tissue: tumor (TUM), stroma (STR), lymphocytic infiltrate (LYM), and necrosis (NEC). 

We compare our method with several  SOTA methods including HistoSegNet \cite{chan2019histosegnet},
SEAM \cite{du2022weakly},
C-CAM \cite{chen2022c},
WSSS-Tissue \cite{han2022multi}, and
PistoSeg \cite{fang2023weakly}, as described in Section~\ref{sec:related}.

\subsection{Settings}
All experiments are conducted with 3 Nvidia RTX 3090 GPUs and environment of PyTorch 1.12.0. Resnet38d is applied for classification part. We choose transformed cross entropy loss for segmentation Loss. Adam optimizer is adopted to optimize the models. We train the model in 10 epoches and the training rate is set to 0.001.


 
\subsection{Results}

\begin{figure}[htbp]
  \centering
  \includegraphics[width=0.8\textwidth]{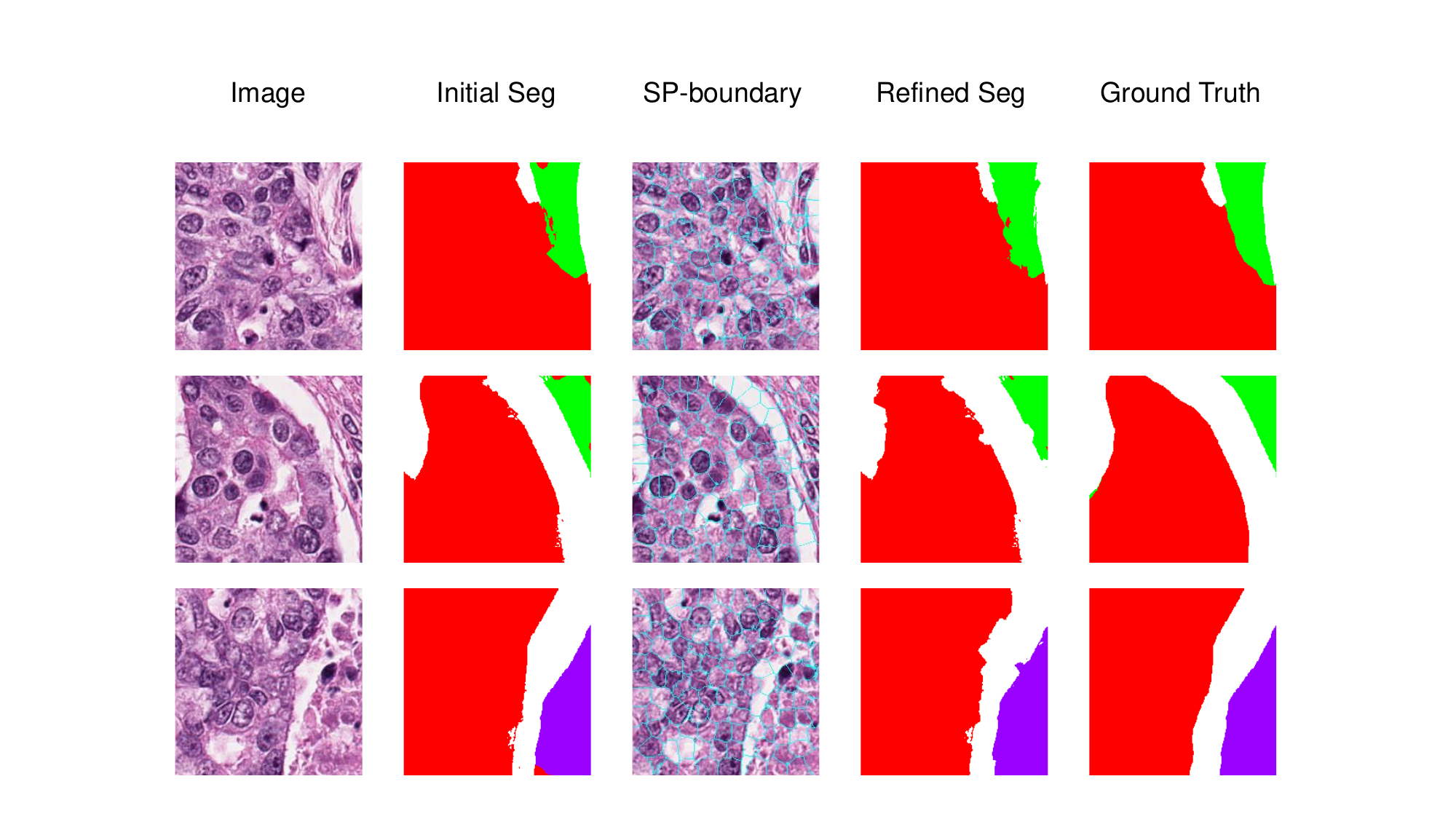}
  \caption{Visualization of superpixel floodfill refinement for BCSS dataset.}
  \label{cmap}
\end{figure}

The results of our method and other SOTA methods are summarized in Table. ~\ref{tab:res-bcss}. Our method achieves the best performance on the overall metric mIoU and performs exceptionally well across all categories. While improving average performance, it ranks the first in NEC and the second in TUM and STR. This demonstrates an excellent balance of our method, even surpassing the relatively strong PistoSeg~\citep{fang2023weakly}. Note that PistoSeg utilizes a pre-trained model with data augmentation while our method does not. Among methods without pre-trained models, our method not only exhibits better balance but also achieves the strongest segmentation capability across subgroups.
We also present the visual results of our algorithm in Fig. ~\ref{cmap}, specifically demonstrating how our method refines the boundaries based on superpixels. Some unnatural outlier blocks in the CAM are eliminated, and the boundaries are better aligned with the visual distribution.

\begin{table}[htbp]
  \centering
  \caption{Comparision of recent methods and our superpixel correction method on BCSS dataset.} 
  \label{tab:res-bcss}  
  {
  \begin{tabular}{l cccc c}
    \toprule
    & \multicolumn{4}{c}{IoU}\\
    \cline{2-5}    
      Method  & TUM &  STR &  LYM & NEC & mIoU \\
    \midrule
    HistoSegNet \cite{chan2019histosegnet} & 0.3314 & 0.4646 & 0.2905 & 0.0191 & 0.2764 \\
    SEAM \cite{du2022weakly} &0.7437 & 0.6216 & 0.5079 & 0.4843 & 0.5894 \\
    C-CAM \cite{chen2022c} & 0.7557 & 0.6796 & 0.3100 & 0.4943 & 0.5599 \\
    WSSS-Tissue \cite{han2022multi} & 0.7798 & 0.7295 & \textit{0.6098} & \textit{0.6687} & 0.6970 \\
    PistoSeg \cite{fang2023weakly} & \textbf{0.8110} & \textbf{0.7504} & \textbf{0.6184} & 0.6422 & \textit{0.7055} \\
    \midrule
    Our method & \textit{0.8082} &  \textit{0.7471} & 0.6048 & \textbf{0.6831} & \textbf{0.7108} \\
    \bottomrule
  \end{tabular}
  }
\end{table}

\section{Discussion}

During the experiments, one point of interest is to determine at which stage our proposed superpixel refinement leads to the greatest improvement in segmentation performance. There are two stages of the image worth refining: one is the pseudo-mask generated at the end of the classification stage, and the other is the final predicted image produced at the segmentation stage. We performed refinement at both stages, and the results reveal that refining the second stage significantly improves segmentation performance, whereas refining the first stage has no effect. A reasonable explanation is that adjusting CAMs too early might result in the lack of information from the image. Since WSSS relies on image-level information, modifying the boundaries in the first stage could lead to the loss of important features, causing the second stage labels to contain fewer effective features. Therefore, adjustments to CAMs need to be done cautiously, and refinement at the right stage contributes to significantly effective improvement.

\bibliographystyle{unsrt}  
\bibliography{main}

\begin{thebibliography}{10}

\bibitem{yang2017suggestive}
Lin Yang, Yizhe Zhang, Jianxu Chen, Siyuan Zhang, and Danny~Z Chen.
\newblock Suggestive annotation: A deep active learning framework for biomedical image segmentation.
\newblock In {\em Medical Image Computing and Computer Assisted Intervention- MICCAI 2017: 20th International Conference, Quebec City, QC, Canada, September 11-13, 2017, Proceedings, Part III 20}, pages 399--407. Springer, 2017.

\bibitem{liu2020semi}
Quande Liu, Lequan Yu, Luyang Luo, Qi~Dou, and Pheng~Ann Heng.
\newblock Semi-supervised medical image classification with relation-driven self-ensembling model.
\newblock {\em IEEE Transactions on Medical Imaging}, 39(11):3429--3440, 2020.

\bibitem{chan2021comprehensive}
Lyndon Chan, Mahdi~S Hosseini, and Konstantinos~N Plataniotis.
\newblock A comprehensive analysis of weakly-supervised semantic segmentation in different image domains.
\newblock {\em International Journal of Computer Vision}, 129(2):361--384, 2021.

\bibitem{han2022wsss4luad}
Chu Han, Xipeng Pan, Lixu Yan, Huan Lin, Bingbing Li, Su~Yao, Shanshan Lv, Zhenwei Shi, Jinhai Mai, Jiatai Lin, et~al.
\newblock Wsss4luad: Grand challenge on weakly-supervised tissue semantic segmentation for lung adenocarcinoma.
\newblock {\em arXiv preprint arXiv:2204.06455}, 2022.

\bibitem{zhou2016learning}
Bolei Zhou, Aditya Khosla, Agata Lapedriza, Aude Oliva, and Antonio Torralba.
\newblock Learning deep features for discriminative localization.
\newblock In {\em Proceedings of the IEEE Conference on Computer Vision and Pattern Recognition}, pages 2921--2929, 2016.

\bibitem{shen2017deep}
Dinggang Shen, Guorong Wu, and Heung-Il Suk.
\newblock Deep learning in medical image analysis.
\newblock {\em Annual Review of Biomedical Engineering}, 19(1):221--248, 2017.

\bibitem{hanna2020whole}
Matthew~G Hanna, Anil Parwani, and Sahussapont~Joseph Sirintrapun.
\newblock Whole slide imaging: technology and applications.
\newblock {\em Advances in Anatomic Pathology}, 27(4):251--259, 2020.

\bibitem{chan2019histosegnet}
Lyndon Chan, Mahdi~S Hosseini, Corwyn Rowsell, Konstantinos~N Plataniotis, and Savvas Damaskinos.
\newblock Histosegnet: Semantic segmentation of histological tissue type in whole slide images.
\newblock In {\em Proceedings of the IEEE/CVF International Conference on Computer Vision}, pages 10662--10671, 2019.

\bibitem{du2022weakly}
Ye~Du, Zehua Fu, Qingjie Liu, and Yunhong Wang.
\newblock Weakly supervised semantic segmentation by pixel-to-prototype contrast.
\newblock In {\em Proceedings of the IEEE/CVF Conference on Computer Vision and Pattern Recognition}, pages 4320--4329, 2022.

\bibitem{chen2022c}
Zhang Chen, Zhiqiang Tian, Jihua Zhu, Ce~Li, and Shaoyi Du.
\newblock C-cam: Causal cam for weakly supervised semantic segmentation on medical image.
\newblock In {\em Proceedings of the IEEE/CVF Conference on Computer Vision and Pattern Recognition}, pages 11676--11685, 2022.

\bibitem{han2022multi}
Chu Han, Jiatai Lin, Jinhai Mai, Yi~Wang, Qingling Zhang, Bingchao Zhao, Xin Chen, Xipeng Pan, Zhenwei Shi, Zeyan Xu, et~al.
\newblock Multi-layer pseudo-supervision for histopathology tissue semantic segmentation using patch-level classification labels.
\newblock {\em Medical Image Analysis}, 80:102487, 2022.

\bibitem{fang2023weakly}
Zijie Fang, Yang Chen, Yifeng Wang, Zhi Wang, Xiangyang Ji, and Yongbing Zhang.
\newblock Weakly-supervised semantic segmentation for histopathology images based on dataset synthesis and feature consistency constraint.
\newblock In {\em Proceedings of the AAAI Conference on Artificial Intelligence}, volume~37, pages 606--613, 2023.

\end{thebibliography}

\end{document}